\pdfoutput=1

\documentclass[11pt]{article}

\usepackage[preprint]{acl}
\usepackage{float}
\usepackage{times}
\usepackage{latexsym}
\usepackage{booktabs}
\usepackage{multirow}
\usepackage{multicol}
\usepackage{cancel}
\usepackage{colortbl} 
\usepackage{xcolor}

\usepackage{amsmath}
\usepackage{longtable}
\usepackage{array}
\usepackage{hhline} 
\usepackage{tabularx}
\usepackage{hyperref}
\usepackage{soul} 
\usepackage{pifont} 
\usepackage{booktabs} 
\usepackage{csquotes} 
\usepackage[utf8]{inputenc}
\usepackage{fontawesome} 
\usepackage{textalpha}  

\definecolor{MyDarkGreen}{RGB}{50, 139, 50} 

\definecolor{MyOrange}{RGB}{255, 140, 0} 
\newcommand{\cmark}{\textcolor{MyDarkGreen}{\ding{51}}} 
\newcommand{\xmark}{\textcolor{red}{\ding{55}}} 

\newcommand{\warn}{\textcolor{MyOrange}{\textbf{?}}} 
\newcommand{\ignore}[1]{} 
\usepackage[T1]{fontenc}

\usepackage[utf8]{inputenc}

\usepackage{microtype}

\usepackage{inconsolata}

\usepackage{graphicx}

\title{Automating Violence Detection and Categorization from Ancient Texts}

  \author{Alhassan Abdelhalim \and Michaela Regneri\\
  Universität Hamburg, Dept. of Computer Science, Hamburg, Germany \\
  \texttt{\{alhassan.abdelhalim, michaela.regneri\}@uni-hamburg.de} 
 }

\begin{document}
\maketitle
\begin{abstract} 
  Violence descriptions in literature offer valuable insights for a wide range of research in the humanities. For historians, depictions of violence are of special interest for analyzing the societal dynamics surrounding large wars and individual conflicts of influential people. Harvesting data for violence research manually is laborious and time-consuming. This study is the first one to evaluate the effectiveness of large language models (LLMs) in identifying violence in ancient texts and categorizing it across multiple dimensions. Our experiments identify LLMs as a valuable tool to scale up the accurate analysis of historical texts and show the effect of fine-tuning and data augmentation, yielding an F1-score of up to 0.93 for violence detection and 0.86 for fine-grained violence categorization.



\end{abstract}

\section{Introduction} \label{sec:introduction}

Violence has been a defining element in human history, influencing cultural values, political structures, and social norms \cite{1130000794906883456, raaflaub2007war, konstan2007emotions}. Understanding its role in shaping ancient civilizations provides valuable insights into societal evolution, power dynamics, and conflict resolution \cite{westbrook2003history, redfield1994nature,bizos2008ethics}. To analyze historical texts for information on violent events, historians have traditionally relied on manual analysis, reading, and annotating vast amounts of text. While manual annotation remains a gold standard for nuanced interpretations, time and labor required for the sheer volume of ancient texts and their linguistic complexities make this approach intractable for exhaustive collections of ancient manuscripts. The rapid growth of digital archives and historical corpora underscores the need for automated methods to assist historians in extracting information more efficiently. 

Large Language Models (LLMs), such as BERT \cite{devlin-etal-2019-bert}, RoBERTa \cite{liu2019roberta}, and GPT \cite{radford2018improving}, have successfully been applied to a wide range of classification tasks, also for scaling annotation of historical texts \cite{electronics13244990}. So far, they have not been used to classify text passages denoting violent events. 

Our research bridges the gap between the hermeneutical processes of historical analysis and the computational methods of natural language processing. We develop and evaluate methodologies that automate the annotation of violence in ancient texts while preserving the depth of understanding traditionally achieved through manual methods. As our gold standard, we use the manually curated ERIS database \cite{riess2015eris}\footnote{\url{https://www.ancientviolence.uni-hamburg.de}}, a large digital collection of violent events from ancient literature. 

We first identify the violent passages contained in ERIS within their original texts using classifiers based on LLMs. Then, we further reproduce some more fine-grained annotations from ERIS, categorizing the violent passages across multiple dimensions: level of violence, contextual background, underlying motives, and long-term consequences.
The results of our study show that LLMs offer a promising solution for extracting violence data. They can expedite the identification of violent events and the extraction of contextual information from ancient texts. With accurate results for a range of classification tasks around violence, LLMs can complement the expertise of historians, allowing them to focus on deeper interpretative tasks rather than the extensive and time-consuming data processing typically required.

In the following, we first give an overview of related work (Sec.~\ref{related}) before we introduce our dataset and methodology (Sec.~\ref{setup}). We then present our results (Sec.~\ref{results}) and discuss their implications (Sec.~\ref{sec:discussion}) before we conclude with a short summary and ideas for future work (Sec.~\ref{conclusion}). Code and data are provided as supplementary material \footnote{\url{https://osf.io/ae835/ }}.

\section{Background \& Related Work}\label{related}
This section provides some background on violence research in history and the digital humanities (Sec.~\ref{related:dighum}). We then introduce large language models and discuss related work concerning LLMs for classification and annotation support (Sec.~\ref{related-LLMs}).

\subsection{Historical Perspectives and Data on Violence}\label{related:dighum}
The meaning of violence is deeply shaped by cultural context, making it a complex phenomenon to define. From a historian's point of view, violence can be defined as \textit{"a physical act, a process in which a human being inflicts harm on another human being via physical strength"} \cite{riess2012performing}.
Violence shaped societal values, legal systems, and social hierarchies in ancient civilizations. Interpersonal violence often reflected concepts of honor, justice, and societal expectations, as reflected in texts like The Iliad \cite{diemkehelden, konstan2007emotions}. Legal codes like Hammurabi’s Lex talionis and Roman law institutionalized violence, balancing societal order and retributive justice \cite{roth1995law, 1130000794906883456}.

Power dynamics frequently used violence as a tool for asserting dominance, with leaders such as Julius Caesar and Augustus consolidating power through both physical and symbolic acts of violence \cite{1130000794950992000, dando2010ides}. Gendered violence highlighted patriarchal structures, as myths and legal frameworks depicted male dominance and societal control \cite{lerner1986creation, pomeroy2011goddesses}. Conflict resolution in ancient texts ranged from violent duels to legal settlements and diplomatic treaties, such as the peace treaty after the Battle of Kadesh \cite{witham2020battle, gagarin1982organization}.

Psychological drivers of violence, such as honor, revenge, and emotional turmoil, are central to narratives like The Iliad and The Oresteia, where cycles of vengeance reflect societal norms and the transition to judicial systems \cite{olson1990stories, Cohen_1986}. Violence in historical accounts, such as Caesar’s assassination, also humanizes figures, exposing vulnerabilities and the socio-political landscapes of their time \cite{tranquillus1962twelve, allen2005alexander}.

Analyzing violence in ancient texts enables researchers to gain insights into societal evolution \cite{westbrook2003history}, comparative legal systems \cite{trigger2003understanding,redfield1994nature}, and the foundation of modern justice \cite{jackson1968evolution,bizos2008ethics,10.7551/mitpress/8041.003.0003}.  Detecting violent instances in ancient texts presents unique challenges due to the implicit and symbolic nature of violence in historical narratives. 

In digital humanities, the study of violence in ancient texts relies on digital resources which provide access to extensive literary and historical collections. In our work, we focus on two of these resources:

\textbf{Perseus}\footnote{\url{http://www.perseus.tufts.edu/}} \cite{smith2000perseus} offers Greek and Roman literature with translations, linguistic annotations, and open-access tools, enabling tasks like text reconstruction  and model training \cite{assael-etal-2019-restoring}. Despite its utility, it faces usability challenges \cite{lang2018review,preece2009perseus}.

\textbf{ERIS} \cite{riess2015eris} is a curated and expanding database of violent depictions in Greek, Roman, and some medieval texts. It includes metadata for bibliographic contexts and details of violent events. We provide a more detailed discussion of ERIS compared to Perseus because ERIS plays a central role in our study and has significant potential for future expansion. In contrast, Perseus, being a widely recognized and extensively documented resource, primarily served as a supplementary source to retrieve non-violent contexts for our dataset. ERIS is further introduced in Sec.~\ref{subsection:ERIS}.
 
\subsection{Large Language Models} \label{related-LLMs}

Large Language Models (LLMs) have revolutionized Natural Language Processing (NLP), enabling advanced text understanding and generation capabilities that were previously unattainable. Built on the architecture of Transformers \cite{vaswani2017attention}, LLMs such as Generative Pre-trained Transformer (GPT) \cite{radford2018improving}, BERT \cite{devlin-etal-2019-bert}, and RoBERTa \cite{liu2019roberta} have set new benchmarks in language modeling and processing tasks.

GPT excells in generative tasks like text completion and translation by leveraging a unidirectional architecture that predicts the next word based on prior context \cite{brown2020language}. In contrast, BERT (Bidirectional Encoder Representations from Transformers) introduced bidirectional context understanding, enabling deeper insights for tasks such as question answering and named entity recognition \cite{devlin-etal-2019-bert}. RoBERTa (Robustly Optimized BERT Pretraining Approach) further refined BERT’s capabilities by using larger datasets for training and optimizing various hyperparameters, which enhances performance across various benchmarks \cite{liu2019roberta}.

These models demonstrate the power of pretraining on vast datasets, capturing linguistic patterns and contextual nuances that generalize across diverse domains. In consequence, LLMs cemented their role as main component for scalable language processing, especially various classification tasks, such as sentiment analysis \cite{bang-etal-2023-multitask}, text categorization \cite{liang2022holistic}, and natural language inference \cite{honovich2022true}. The possibility to fine-tune such pre-trained models to small domains makes them a versatile tool also for uncommon data like ancient texts: 
They have already been used for scaling up annotation of historical data \cite{electronics13244990}, and for hate speech detection \cite{Mathew_Saha_Yimam_Biemann_Goyal_Mukherjee_2021}.
Both tasks have goals close to our objective of extracting and categorizing violence from ancient texts. Our method is developed to scale the annotation of violent events in ancient texts, and we are also concerned with textually manifested ferocity. Our contribution extends previous approaches in that we use annotation methods for violent texts and that our data contains descriptions of violence rather than verbal assaults, as in hate speech. To the best of our knowledge, we present the first study that automatically extracts and annotates violence from historical text data.

\section{Data and Experimental Setup} \label{setup}

In this section, we explain ERIS as the basis for our experiments (Sec.~\ref{subsection:ERIS}), how we set up the experiments for violence detection (Sec.~\ref{subsection:violence}) and violence categorization (Sec.~\ref{subsection:categorization}), and introduce the evaluation metrics used for both tasks (Sec.~\ref{subsection:evaluation}). 
\subsection{Data: The ERIS Database}
\label{subsection:ERIS}
ERIS \cite{riess2015eris} is a manually curated and continuously growing database containing depictions of violence from Greek, Roman and some medieval texts, including references to violence from Herodian, Plutarch, Tacitus, Thucydides and Xenophon. Each text passage is annotated with metadata, denoting the bibliographic contexts as well as details on the violent event. Among other labels, it categorizes violent acts by context, motives, and social factors. It also provides metadata as timestamps and geographical coordinates, supporting advanced filtering and geospatial analysis. ERIS emphasizes sociological dimensions of violence, enabling a deeper understanding of its impacts across time and regions. Most notably, ERIS contains links to the Perseus database to match violence passages to their original texts. 
At the time of writing this paper, ERIS contained 3,252 entries spanning various time periods, starting from Archaic Greece in the 7th century BCE to the Salian period in the 11th century AD. 

\begin{figure}[H]
    \centering
    
    \small 

    \begin{tabular}{|p{2cm}|p{4.5cm}|}  
        \hline
                    \rowcolor[gray]{0.8} \textbf{Attribute} & \textbf{Details} \\
        \hhline{==}  
        \textbf{Related \newline Conflict} & Wars of Alexander The Great \\
        \hhline{==}  
          \textbf{Perpetrator} & 
        \emph{Name:} Alexander III the Great \newline
        \emph{Age:} Adult \newline
        \emph{Activity:} Monarch/Ruler \newline
        \emph{Origin:} Macedonian \\
        \hhline{==}  

          \textbf{Victim} & 
        \emph{Name:} Cleitus the Black \newline
        \emph{Age:} Adult \newline
        \emph{Direct Consequence:} death \newline
        \emph{Origin:} Macedonian \\
        \hhline{==}  

          \textbf{Third Party (Person)} &
 \emph{Name:} Aristophanes \newline
        \emph{Age:} Adult \newline
                \emph{Activity:} Soldier \\
                
        \textbf{Third Party (Group)} &
        Friends of Alexander III \newline \emph{Origin}: Mixed \newline \emph{Age}: mixed \newline \emph{Activity}: commander/general \\ 
        \hhline{==}  

        \textbf{Source} & Plutarch, Alexander 51.5 \\
        \hhline{==}  
        \textbf{Year} & 328 B.C. \\
        \hhline{==}  
        \textbf{Location} & Maracanda (Samarkand) \\
        \hhline{==}  
         \textbf{Time Period} &Hellenistic Greece \\
        \hhline{==}  
        \textbf{Level} & Interpersonal \\
        \hhline{==}  
        \textbf{Context} & entertaining \\
        \hhline{==}  
        \textbf{Motivation} & emotional \\
        \hhline{==}  
        \textbf{Weapon} & Spear \\
        \hhline{==}  

        \textbf{Original Text} & 
        \textit{"οὕτω δὴ λαβὼν παρά τινος τῶν δορυφόρων Ἀλέξανδρος αἰχμήν ἀπαντῶντα τὸν Κλεῖτον αὐτῷ καὶ παράγοντα τὸ πρὸ τῆς θύρας παρακάλυμμα διελαύνει."} \\
        \hhline{==}
        \textbf{Translation } & 
        \textit{"And so, at last, Alexander seized a spear from one of his guards, met Cleitus as he was drawing aside the curtain before the door, and ran him through."} \\
           \hhline{==}  
           \textbf{Remark} & \emph{perpetrator}: Alexander is shocked by his deed and tries to kill himself. This is mentioned in 51.6. \newline
\emph{thirdperson}: The presence of these persons is mentioned in 51.1-4 and 51.6.\\
 
        \hline  
    \end{tabular}
    \caption{An entry from ERIS titled : 
Alexander kills Cleitus with a spear.}\label{eris:example}
\end{figure}

Figure \ref{eris:example} shows an example entry from ERIS. 
Each entry includes metadata such as title, source references, historical period, and century, as well as detailed classifications of violence level, context, motive, weapon, consequences, and method of execution. Additionally, it provides temporal and situational context, including date, season, month, and duration, along with references to the primary text sources. Some of the attributes also refer to information not contained in the text passage, here noted as \emph{Remark}. 
ERIS mostly contains Greek and Roman literature, along with English translations. Our work is based on the ERIS content from biographies of Plutarch, an ancient Greek writer. We work with the English translations of the original texts.

\subsection{Violence Detection}\label{subsection:violence}
In our first experiment, we perform a binary classification task to detect instances of violence (and distinguish them from non-violent passages) in ancient texts. For classification, we compare the plain pre-trained models with fine-tuned LLMs. 

As ERIS contains only violent passages, we additionally need comparable non-violent examples to train our model. To obtain those, we retrieve the context of the violent passages from ERIS by re-connecting them to their source texts. Then we train LLMs to distinguish violent from non-violent passages. As a baseline, we also use the ChatGPT-API to simulate an annotator that works with the support of ChatGPT and compare the results.

\subsubsection*{Data Pre-processing}\label{preprocess}
To obtain data that we can use for training and testing, we need to amend the ERIS data with nonviolent examples. Our core idea is to retrieve data from the original texts the ERIS passages were extracted from and use sentences that are not labeled in ERIS as nonviolent data. Because this requires us to have digital access to the respective original texts, we restrict this experiment to ERIS samples from Plutarch's biographies, which have digital links to their source text in the Perseus database.

For each violent passage from ERIS, the full sections from which these excerpts were derived were retrieved. Any paragraph not explicitly marked as violent in ERIS was treated as non-violent, forming the negative examples for the dataset, resulting in a final dataset of 461 violent and 2103 non-violent texts extracted from 13 different Plutarch books. We assume that for any book that is completely annotated for ERIS, each text part that is not contained in ERIS does only contain non-violent text. This assumption might not always hold, because annotators could have missed some passages. We discuss future assessment of this in the \hyperref[sec:limitation]{Limitations }section.

\begin{figure}[H]
\hspace{2mm} 
\centerline{\includegraphics[width=0.47\textwidth]{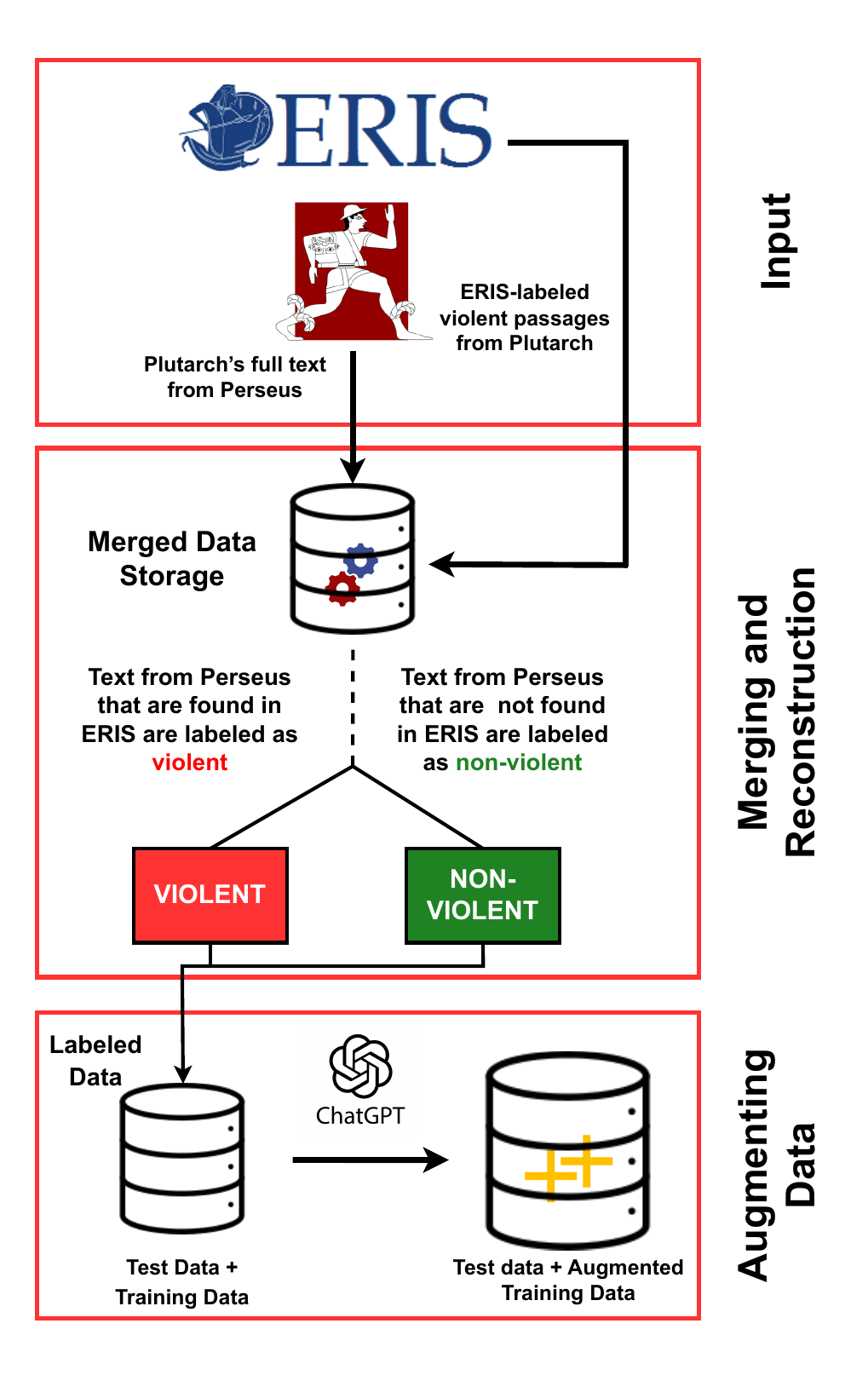}}
	 {\caption{Data Preprocessing Pipeline for Violence Detection.}\label{fig:preprocessing_snippit}}
\end{figure}

As a held-out test set for evaluation, 500 texts (371 non-violent and 129 Violent) distributed evenly across the 13 Plutarch books were selected. The remaining passages were used later on for training, fine-tuning, and data augmentation.

Because the Plutarch vitae constitute a comparably small dataset, we want to explore the impact of data augmentation with LLMs, for which we used the OpenAI API. For each entry in the training dataset, we generated three paraphrased versions and thus quadrupled the training data size. We retained 6928 non-violent texts and 1328 violent texts. Then we tested the augmented models on the 500 sentences reserved for evaluation to quantify the influence of augmentation. Figure \ref{example:augment} shows an exemplary sentence from ERIS and the three paraphrases obtained using the OpenAI API.
\begin{figure}[h]
    \centering
    \footnotesize 
    \renewcommand{\arraystretch}{1.3} 
    \rowcolors{2}{green!15}{white} 
    \begin{tabular}{|>{\centering\arraybackslash}m{1.2cm}|m{5.8cm}|}  
        \hline
        \rowcolor{orange!20} \textbf{Original Sentence} & 
        \textit{As a result of this battle, the Greeks could not only harry the country of the King without fear, but had the satisfaction of seeing due punishment inflicted upon Tisaphernes, an abominable man, and most hateful to the Greek race.  \textbf{[Plut. Ages. 10.3]} }\\
        \hhline{==}  
        \textbf{Para-\newline phrase 1} & 
        \textit{Consequently, the Greek army found themselves able to pillage the king's realm without trepidation, while also deriving satisfaction from the punishment dealt to Tisaphernes, a detestable individual who was particularly reviled by their people.} \\
        \hhline{==}  
        \textbf{Para-\newline phrase 2} & 
        \textit{This battle enabled the Greeks to wreak havoc across the king's lands unimpeded, while they also relished the sight of Tisaphernes, an odious person hated by the Greek civilization, finally facing the consequences of his actions.} \\
        \hhline{==}  
        \textbf{Para-\newline phrase 3} & 
        \textit{Following this conflict, the Hellenic forces were not only able to raid the territory of the monarch freely but also took pleasure in witnessing Tisaphernes, a loathsome figure deeply despised by the Greeks, receive just retribution.} \\
        \hline  
    \end{tabular}
    \caption{An example of our text augmentation. This approach effectively quadriples the training data.} \label{example:augment}
\end{figure}

\subsubsection*{Model Selection}
For violence detection, we evaluated BERT-large and RoBERTa-large as LLMs for direct classification. Both models were tested first as-is and then with fine-tuning to the historical data. As an additional model that simulates annotation with the support of ChatGPT, we used the GPT-4o mini API with a specifically designed prompt that outputs the classifications. We provide the prompt in \autoref{chatgptprompt}.


\subsection{Categorizing Violent Events}\label{subsection:categorization}

In our second experiment, we automatically apply a more fine-grained annotation of violent texts, aiming to reproduce some ERIS annotations. In contrast to the first experiment, we use the full ERIS database as our source data. Thus, our input contains a wider variety of source texts than the violence classification (which was restricted to Plutarch biographies) and only texts that are manually labeled as violent. For this experiment, data augmentation was not suitable because we would have to augment the fine-grained annotation from ERIS as well. 

With the ERIS passages, we use a multi-class classification approach across four key dimensions from the ERIS annotations:

\begin{itemize}
    \item \textbf{Level of Violence:} Classifies instances of violence into four categories: interpersonal (conflict between individuals), intrapersonal (self-harm), intersocial (conflicts between groups, like wars), and intrasocial (conflicts within a societal group). They highlight the relational context of the events.
    
    \item \textbf{Context:} Contains 25 categories of the setting in which the violence occurred, with various political, military, and social contexts.

    \item \textbf{Motive:} 13 different classes for the underlying reasons for violent actions, distinguishing between tactical/strategic goals, political ambitions, adherence to authority, emotional impulses, and economic motives.

    \item \textbf{Long-Term Consequences:} The most fine-grained label with 38  outcomes of violent events, including social disruption, political changes, and personal impacts.
\end{itemize}

We split the dataset into 80\% for training/validation and 20\% for testing. Some (5) classes with very few instances do not occur in the randomly assigned test split. We fine-tuned and evaluated one BERT and one RoBERTa model per dimension.

\subsection{Evaluation Metrics}
\label{subsection:evaluation}
For both experiments, we measure the performance of all models using the standard evaluation metrics precision, recall and F1 score. 

Given that $TP$, $FP$, $FN$, $TN$ are the True Positives, False Positives, False Negatives, and True Negatives respectively, key metrics are defined as follows:

\begin{equation}
    \textbf{Precision:} \ \quad P = \frac{TP}{TP + FP}
\end{equation}
    Precision measures the proportion of correct positive predictions.
\begin{equation}
    \textbf{Recall:} \quad R = \frac{TP}{TP + FN}
\end{equation}
    Recall measures the proportion of actual positives that are correctly identified.
\begin{equation}
    \textbf{F1 Score:} \quad F_1 = \frac{2*P*R}{P + R} 
\end{equation}
The F1 score is the harmonic mean of precision and recall, which is sensitive to disparities between them. This property ensures that the F1 score is low if either precision or recall is low, accurately reflecting the model's overall performance. \\

We also provide two baselines: majority and random. A \textbf{majority baseline} represents a trivial classifier that only predicts the majority class $c_{\text{majority}}$.  Given class probabilities $P(c_k)$ for $K$ classes, a class $c_k$ is predicted with: \(
\hat{y} = c_{\text{majority}}, \quad \forall x \in X
\). \\

A \textbf{random baseline} assigns labels based on class probabilities \( p_i = \frac{\#C_i}{X} \). The expected probability of making a correct prediction is given by \(\sum_{i=1}^{N} p_i^2\). This represents the probability of randomly guessing the correct label, serving as a lower-bound benchmark for classifiers.

\section{Results}\label{results}
In this section we provide our results for violence detection (Sec.~\ref{results: violence-detection}) and violence categorization (Sec.~\ref{results:violence_categorization}).
\subsection{Violence Detection} \label{results: violence-detection}

Our results are summarized in \autoref{Table_Merged_Results1}. Overall, BERT with augmentation and fine-tuning performs best for our task.
Fine-tuning enhanced the results drastically. The fine-tuned BERT and RoBERTa-large yielded an F1-score of 0.83 and 0.87, effectively capturing violent instances. Both provided competitive results.

\begin{table*}[ht]
    \centering
\resizebox{0.8\textwidth}{!}{ 
    \begin{tabular}{llccccc}
        \toprule
        & \textbf{Model} & \textbf{Precision} & \textbf{Recall} & \textbf{F1-Score} & \textbf{Support} \\
                \midrule
        \textbf{Violent}     & GPT-4o mini   & 0.69 & 0.74 & 0.71 & 129 \\
                    & BERT [as-is] & 0.25 & 0.97 & 0.40 & 129 \\
                    & BERT [fine-tuned]         & 0.88 & 0.78 & 0.83 & 129 \\
                    & BERT  [fine-tuned and augmented] & 0.87 & 0.99 & \textbf{0.93} & 129 \\
                    & RoBERTa [as-is] & 0.00 & 0.0 & 0.00 & 129 \\
                    & RoBERTa  [fine-tuned]     & 0.89 & 0.86 & 0.87 & 129 \\
                    
                    & RoBERTa  [fine-tuned and augmented] & 0.82 & 0.99 & \textbf{0.90} & 129 \\
        \midrule
        \textbf{Non-Violent} & GPT-4o mini   & 0.91 & 0.88 & 0.89 & 371 \\
                    & BERT [as-is]       & 0.00 & 0.00 & 0.00 & 371 \\
                    & BERT [fine-tuned]         & 0.93 & 0.96 & 0.94 & 371 \\
                    & BERT  [fine-tuned and augmented] & 1.00 & 0.95 & \textbf{0.97} & 371 \\
                    & RoBERTa [as-is] & 0.74 & 1.0 & 0.85 & 371 \\
                    & RoBERTa [fine-tuned]      & 0.95 & 0.96 & 0.96 & 371 \\
                    
                    & RoBERTa [fine-tuned and augmented] & 1.00 & 0.92 & \textbf{0.96} & 371 \\

        \midrule
        \textbf{Overall} & GPT-4o mini   & 0.69 & 0.74 & 0.71 & 500 \\
                         & BERT [as-is] & 0.25 & 0.97 & 0.40 & 500 \\
                         & BERT [fine-tuned]   & 0.88 & 0.78 & 0.83 & 500 \\
                         & BERT [fine-tuned and augmented] & 0.87 & 0.99 &  \textbf{ 0.93} & 500 \\
                         & RoBERTa [as-is] & 0.00 & 0.00 & 0.00 & 500 \\
                         & RoBERTa   [fine-tuned]     & 0.89 & 0.86 & 0.87 & 500 \\
                         
                         & RoBERTa [fine-tuned and augmented] & 0.82 & 0.99 & \textbf{0.90*} & 500 \\
        \midrule
        \textbf{Baselines (overall)} & Majority (all non-violent) & {0.74} & {0.74} & {0.74} &  500 \\
                         & Random   & 0.61 & 0.61 & 0.61 & 500 \\
        \bottomrule
    \end{tabular} 
    }
    \caption{Violence detection performance across different models, evaluated per class (Non-Violent and Violent). Support indicates the number of instances in each class of the test set. (*) marks an insignificant difference.}
    \label{Table_Merged_Results1}
\end{table*}

Applying data augmentation enhanced the performance of both models. In particular, it vastly enhanced recall for all models, which is of particular interest for supporting annotators: The most common mistake when extracting violent passages manually is to miss them in the text. Having a preprocessor with high recall (maybe compromising with lower precision) can perfectly complement the precise human annotation because it is much faster to sort out falsely selected violent passages than to re-read the whole source text to retrieve missed but relevant paragraphs.

For F1, data augmentation made only a significant difference for BERT (p < 0.05 using McNemar's test), but not for RoBERTa.

Our simulated zero-shot GPT annotator implemented with the general-purpose GPT-4o mini model attained an F1-score of 0.71 but struggled with non-violent instances. We attribute this to the lack of fine-tuning in ChatGPT, which is supported by both our results and many other studies that measure the importance of domain-specific fine-tuning for better classification \cite{rietzler-etal-2020-adapt, rostam2024fine, liu-etal-2024-panda}. We also evaluated the larger GPT-4o model exploratively, which is approximately 16 times more expensive than the GPT-4o mini version. Despite the increased computational cost, GPT-4o offered only marginal improvements on our test data in the F1 score (0.5), indicating limited practical advantage for this task. We thus continued using the GPT-4o mini model.

\subsection{Violence Categorization} \label{results:violence_categorization}
For categorization, we used fine-tuned BERT-large and RoBERTa-large. An overview of the results is shown in \autoref{Table_Consolidated_Comparison_with_Baselines}. We report the averages over all instances, which amounts to weighted averages over the individual classes. A detailed breakdown by individual labels is given in \autoref{exp2_breakdown}. We generally achieve promising results with an F1 score of 0.8, even for the most fine-grained category (long-term consequences with 37 classes). As for violence identification, BERT shows a slightly better performance than RoBERTa.

For identifying the \emph{violence level}, the models performed best in classifying interpersonal and intersocial violence, achieving high precision and recall. However, intrapersonal violence posed challenges due to its lower representation and the subtle contextual understanding required.

For \emph{context}, F1 is still comparably high given the complexity of the task with 23 classes. Looking at the details, we find that the model effectively identified broad categories like "War/Military Campaign" and "Battle" but struggled with nuanced distinctions between similar contexts, such as large-scale campaigns versus single combat.

\begin{table*}[hbt!]
    \centering
    \small
    \begin{tabular}{llccc|ccc|cc}
        \toprule
        & & \multicolumn{3}{c}{\textbf{RoBERTa}} 
        &  \multicolumn{3}{c}{\textbf{BERT}} 
        &  \multicolumn{2}{c}{\textbf{Baselines }} \\
        \cmidrule(lr){3-5} \cmidrule(lr){6-8} \cmidrule(lr){9-10}
        \textbf{Dimension} & \textbf{Classes} 
        & \textbf{Precision} & \textbf{Recall} & \textbf{F1-Score} 
        & \textbf{Precision} & \textbf{Recall} & \textbf{F1-Score} 
        & \textbf{Majority} & \textbf{Random} \\
        \midrule
        \hyperref[Table_Level_Comparison_with_Baselines]{Level}                  & 4 & 0.95 & 0.95 & \textbf{0.95} & 0.93 & 0.93 & 0.93 & 0.67 & 0.49 \\
        \hyperref[Table_Context_Comparison_with_Baselines]{Context}                & 23 & 0.86 & 0.85 & 0.84 & 0.86 & 0.86 & \textbf{0.85} & 0.33 & 0.16 \\
        \hyperref[Table_Motive_Comparison_with_Baselines]{Motive}                 & 12 & 0.86 & 0.85 & 0.85 & 0.86 & 0.86 & \textbf{0.86} & 0.35 & 0.20 \\
        \hyperref[Table_Long_Term_Comparison_with_Baselines]{Consequences} & 37 & 0.82 & 0.80 & 0.80 & 0.82 & 0.81 & \textbf{0.81} & 0.36 & 0.16 \\
        \bottomrule
    \end{tabular} 
    \caption{Overall Violence Categorization Results. Breakdowns by label are provided in \autoref{exp2_breakdown}}
    \label{Table_Consolidated_Comparison_with_Baselines}
\end{table*}

Distinguishing \emph{motives} works with similar accuracy. Again, the model performed well in identifying broad categories like "Tactical/Strategical" and "Political" but struggled with nuanced or less frequent categories such as "Emotional" and "Ambition". Overlaps between motives like "Political", "Following Orders", and "Tactical/Strategical" led to misclassifications.

Finding the \emph{Long-Term Consequences} was the most challenging task with 37 different classes. The model excelled in identifying concrete categories like "Destruction/Devastation" and "Victory," which are frequently referenced in historical texts. However, categories with fewer examples, such as "Exile" and "Coronation," proved challenging, resulting in lower precision and recall. The abstract nature of some consequences, like political changes or psychological impacts, added complexity to classification.

\section{Discussion}\label{sec:discussion}

The experiments demonstrated the potential of fine-tuned large language models (LLMs) in detecting and classifying violence in ancient texts. Our evaluation demonstrates the models’ strengths in violence classification, with an F1-score of up to 0.93. In manual classification recall is often the problem due to  implicit or symbolic violence, ambiguous wording, and a bias toward precision, leading to missed instances. Our finetuned and augmented models achieve a high recall, showing that LLMs can mitigate blind spots that humans miss. However, challenges like class imbalance, conceptual overlap, and abstract categories in multi-class tasks revealed areas for improvement.

For violence categorization, our approach excelled in well-represented and concrete categories, such as “Victory” and “Destruction,” but struggled with abstract or underrepresented categories like “Intrapersonal Violence” or “Exile”. Conceptual overlaps, such as between “Political” and “Tactical” motives, also led to misclassifications.

From the perspective of historians, choosing between fine-tuned models, tools like ChatGPT, or manual annotation depends on specific project needs. We provide an overview over the specific features to be considered for applying fine-tuned LLMs and ChatGPT (either via user interface or via API) in \autoref{tab:fine-tuning_vs_api}. Fine-tuned LLMs excel in structured, large-scale tasks where efficiency and consistency are paramount, offering rapid processing capabilities that can save months of manual labor. ChatGPT, while versatile and user-friendly, lacks domain-specific fine-tuning, making it less reliable for specialized classifications but valuable for exploratory tasks or initial insights. Manual annotation remains irreplaceable for complex interpretative work, especially in ambiguous cases requiring deep historical expertise. A hybrid approach, where LLMs handle bulk annotation and historians validate edge cases, offers an optimal balance between efficiency and precision. 
\begin{table}[h!]
    \centering
    \small
    \renewcommand{\arraystretch}{1.5} 
    \setlength{\tabcolsep}{3pt} 
    \begin{tabular}{@{}p{0.6\linewidth}cc@{}} 
        \toprule
        \textbf{Criteria}                   & \textbf{LLM Finetuning} & \textbf{API} \\
        \midrule
        Highly specialized task            & \cmark                     & \xmark             \\
        Requires extensive labeled data    & \cmark                     & \xmark             \\ 
        Cost-effective for small tasks     & \xmark                     & \cmark             \\
        Faster deployment                  & \xmark                     & \cmark             \\
        Full control over architecture     & \cmark                     & \xmark             \\
        Local dependency     & \cmark                     & \xmark             \\
        inference speed& \cmark                     & \warn             \\
        Suitable for dynamic scaling       & \xmark                     & \cmark             \\
        Ongoing model maintenance          & \cmark                     & \xmark             \\
        Scalability& \xmark                     & \cmark             \\

        Convenience / Usable across devices  & \xmark                     & \cmark             \\
        Ongoing Maintenance / feedback & \xmark                     & \cmark             \\

        Ethical considerations & \warn                     & \cmark             \\
        \bottomrule
        \bottomrule
    \end{tabular}
    \caption{Pros and cons of fine-tuning LLMs vs. zero-shot approach through pre-trained OpenAI APIs}
    \label{tab:fine-tuning_vs_api}
\end{table}

Convenience and usability are also to be considered when choosing between fine-tuning LLMs or directly using APIs. Fine-tuned models require technical expertise for setup and training but deliver streamlined workflows once operational. ChatGPT, with its accessible API and conversational interface, is more user-friendly and easy to use since it can be conveniently used in tablets or mobile phones. However, it lacks the tailored accuracy of fine-tuned models.while manual annotation is intellectually robust, it is resource-intensive and impractical for large datasets. Integrating intuitive interfaces with fine-tuned models could enhance their usability, encouraging broader adoption among non-technical users. 

Inference speed varies between fine-tuned models and API-based solutions. Fine-tuned models offer lower latency but require dedicated hardware, while API-based models provide scalability but introduce network latency and rate limits. Fine-tuning is preferable for low-latency applications, while APIs offer scalability and ease of use.

Ongoing model maintenance refers to the continuous process of monitoring, updating, and retraining fine-tuned LLMs to maintain their performance and adapt to evolving data distributions or task requirements. When practitioners fine-tune their own models, they bear the responsibility for performance monitoring, infrastructure management, and regular model updates to ensure accuracy and relevance over time.

Ethical and bias considerations differ significantly between fine-tuned LLMs and API-based solutions. Pre-trained APIs are typically pre-moderated, incorporating safeguards to filter harmful or biased outputs. On the other hand, fine-tuned models require custom mitigation strategies \cite{jin-etal-2021-transferability, garimella-etal-2022-demographic}, which can either reduce or amplify biases, depending on dataset quality and training methods. Fine-tuning allows for domain-specific alignment but poses risks if ethical oversight is inadequate.

The implications of this research extend beyond ancient texts, offering valuable insights for analyzing contemporary violence depictions, addressing modern datasets such as media reports, social media content, or legal documents. Adapting the models to contemporary datasets would require adjustments to account for different linguistic styles, cultural contexts, and evolving definitions of violence, presenting an exciting avenue for interdisciplinary research.

A significant gap lies in automating the identification of abstract or highly contextual categories, such as psychological impacts or symbolic violence. Achieving this would require expanding datasets, understanding abstractions in LLMs \cite{regneri2024detecting}, incorporating knowledge bases \cite{wang2024kblam}, and exploring advanced techniques like retrieval augmented generation (RAG) \cite{chen2024benchmarking}. Developing dynamic models that can learn from continuous expert feedback through techniques like reinforcement learning from human feedback (RLHF) could also bridge this gap \cite{kaufmann2023challenges}.

\section{Conclusion}\label{conclusion}

In this work, we proposed a framework for automating the classification and categorization of violent ancient texts using LLMs. Our two main contributions are the development of models capable of accurately classifying violent sentences,  and employing these models to automate the process of fine-grained violence categorization. In both cases, we showed the effect of fine-tuning the models. For violence detection, we also showed that data augmentation drastically enhances recall, which is the most important measure for supporting manual annotation. Our results can enable historians to accomplish tasks that previously required months or years in minutes. To the best of our knowledge, we are also the first to utilize the OpenAI API to classify violent ancient historical texts and compare its performance against other pre-trained models. Our findings underscore the potential of LLMs to automate labor-intensive tasks and pave the way for large-scale text analysis in historical research. While fine-tuned LLMs provide structured and efficient classification, ChatGPT remains useful for exploratory tasks, and manual annotation retains its importance in complex interpretative work.

Challenges remain, particularly with underrepresented classes and computational constraints. Exploring larger models could enhance contextual understanding while maintaining runtime performance. Future work in close collaboration with historians could help resolve ambiguous cases that even human experts find difficult to classify. A hybrid approach integrating automated classification with expert validation would maximize both efficiency and accuracy. Additionally, incorporating surrounding textual context instead of analyzing passages in isolation could further enhance classification performance. Our methods also offer potential for extending the ERIS database to annotate and include texts from more recent historical periods. Adapting the models to contemporary datasets would require adjustments for linguistic style, cultural contexts, and evolving definitions of violence, presenting exciting opportunities for interdisciplinary research.

\section*{Limitations} \label{sec:limitation}
Our study shows a promising approach to scaling up the annotation of violent events in ancient texts. While delivering accurate results in our experiments, we acknowledge several limitations rooted in the dataset, the methodology and the experimental coverage. 

\paragraph{Dataset and annotation:} ERIS is a well-curated dataset and contains the largest amount of manually annotated violent text passages from historical texts. However, this dataset also has its limits: First, for a machine learning approach the number of examples is still comparably small. Second, it only contains historical data from ancient texts as well as some medieval texts. While we assume that our approach would be applicable (possibly after more fine-tuning) to other texts, too, we cannot evaluate it with the given data. Further, ERIS does not contain information on inter-rater agreement, so we do not have a manual comparison stating how complex the task is for humans. We also do not have a detailed account on the amount of time it takes to annotate the violence passages manually. What we do know is that it strongly depends on the annotator, and that manual efforts are, overall, not easy to scale.
\paragraph{Methodology and Experiments:} Given the limits of the database, our experiments have further limitations added. First of all, we only operate on translations rather than original texts. This might be a restriction for both text understanding and scaling the methods to texts for which no translations are available. Currently, this mirrors the manual annotation process, because annotators with knowledge of Latin or Ancient Greek are hard to find, so most of ERIS is annotated using the translations.

For violence classification, we only used the texts available in the Perseus database, because we needed to extend the ERIS data with comparable passages that do not contain violent data. Like this, the violence classification does not contain the whole ERIS database, especially not the medieval texts. While we are convinced that our results can still carry over to other epochs and text sorts, our experiments do not prove this as of yet. 

Some accuracy in the fine-grained violence categorization is lost in the automated annotation, which is partly due to the ambiguity within the texts, and partly due to the challenging fine-grained taxonomy in ERIS. It is up to future work to decide whether the actual annotation guidelines and the categories need to be adapted or whether the methodology should account for this. To make this distinction, more detailed analysis and data on inter-annotator agreement would be needed (see above).

Weighted averages were chosen to reflect overall model performance effectively, particularly given the significant imbalance between class sizes. However, this method inherently favors dominant classes and can obscure weaker results in less frequent categories. A more balanced approach should be considered, potentially involving class-based weighting or specialized metrics to ensure accurate representation across all classes.

Further, we only used four of the fine-grained ERIS categorizations for annotations. We did not do further categorization and information extraction to simulate a complete annotation of an ERIS entry. While we think that some categories are straight-forward to apply (like the identification of the weapon), others might be impossible for a model to guess, because they are not contained in the violent passages (like geographical data or sometimes the actors). In order to do this comprehensive annotation automatically, we would have to implement a different classification approach that takes the context of the violent text passages into account. We leave this experiment for future work.

\section*{Ethics Statement}
We provide an experiment that helps to classify violent text passages, primarily in ancient texts. We did not use or produce any sensitive data during those experiments. We do see the potential for our method to be applied for the common good, especially when adapted to contemporary data. Like other studies on hate speech have shown, the automated detection of harmful content can support the automated analysis of the media with the aim of  \emph{protecting vulnerable groups}.

While the methodology presented in this work is primarily intended for academic and educational purposes, we recognize the potential \emph{misuse} of AI technologies in misrepresenting historical data when applied without supervision. A misclassification of violent text or a blind reliance on the comprehensiveness of the method can lead to unwanted mistakes in the aforementioned protective purposes. Like most statements here, this applies to basically all automation methods and needs to be mediated accordingly.

Bearing in mind the general societal awareness of jobs being automatized, our work explicitly encourages the \emph{responsible use of AI in humanities research}. Our models are designed to complement human expertise, ensuring that tedious workload is alleviated, which might be especially welcome in the case of violent texts.  Like all automation approaches, this aims at scaling in terms of data set size rather than replacing analysis depth. This allows historians to focus on deeper interpretative analyses, fostering a collaborative approach between human expertise and machine learning.

\section*{Acknowledgments}
The authors wish to thank Werner Rieß and Justine Diemke for providing access to the ERIS database and offering valuable feedback on results and potential research directions. Special thanks to Sri Gowry Sritharan for her assistance in extracting non-violent examples from Perseus. Additionally, the authors extend their gratitude to Sören Laue, Hanna Herasimchyk, Lennart Bengtson, and Mostafa Kotb for their technical and methodological advice and for proofreading and improving the manuscript. We also thank the three anonymous reviewers for their helpful comments. All remaining errors are, of course, our own.
\bibliography{bib}

\clearpage 

\appendix
\onecolumn

\section{Code and Data} \label{sup:code_data}
We provide the training data for both tasks as well as the code, downloadable under
\begin{center}
\url{https://osf.io/ae835/}
\end{center}

\subsubsection*{Violence detection}
The folder \verb|1_violence_detection| contains the training and test data for the violence identification task. The sentences are a subset of ERIS extended with their original contexts extracted from the Perseus database. We provide both the original dataset and the augmented dataset used for training. The annotation contains the source as noted in Perseus (book, chapter, and section), the passage text, and the violence annotation (1 for violent, 0 for non-violent).
\subsubsection*{Violence categorization}
The folder \verb|2_violence_categorization| contains a condensed version of the ERIS database, including the text passage with the four annotation dimensions we used for classification. To reproduce our training and test data, please use the code we provide.
\subsubsection*{Code}
We provide two Jupyter notebooks (\verb|violence_detection.ipynb| and \verb|violence_categorization.ipynb|) to reproduce our data preprocessing, model training, and evaluation for both tasks. 

\section{GPT-4o mini Testing Prompt}
\label{chatgptprompt}

\begin{quote}
\itshape
You are a historian that classifies historical texts into violent or non-violent based on the provided examples. The following principles apply to the classification of violent acts:
\begin{itemize}
    \item Arrests of people and banishments are initially recorded as acts of violence and discussed with the team before being activated.
    \item Fictional narratives, such as the conquest of Troy, are included.
    \item Establishment of colonies, verbal violence (insults), and damage to property (including fires in buildings, etc.) are excluded.
\end{itemize}

Your task is to classify each passage based on the criteria above. Respond with only \textbf{[VIOLENT]} or \textbf{[NON-VIOLENT]} for each classification.
\end{quote}

\section{GPT-4o mini Augmentation Prompt} \label{chatgptaugmentationprompt}
\begin{quote}
\itshape
You are a historian that wants to paraphrase sentences to create new ones for enhancing your dataset. Generate three different ways to rewrite the following sentence while keeping the same meaning. Important to note that you are not allowed to change context, motive or consequences.
\end{quote}

\section{Detailed Breakdown for the Violence Categorization results}
\label{exp2_breakdown}
These are the extended results for \autoref{Table_Consolidated_Comparison_with_Baselines}.

\begin{table*}[http]
    \centering
    \scriptsize
    \begin{tabular}{lcccc|cccc}
        \toprule
        & \multicolumn{4}{c}{\textbf{RoBERTa Results}} & \multicolumn{4}{c}{\textbf{BERT Results}} \\
        \cmidrule(lr){2-5} \cmidrule(lr){6-9}
        & \textbf{Precision} & \textbf{Recall} & \textbf{F1-Score} & \textbf{Support} 
        & \textbf{Precision} & \textbf{Recall} & \textbf{F1-Score} & \textbf{Support} \\
        \midrule
        Interpersonal & 0.92 & 0.91 & 0.91 & 96 & 0.93 & 0.88 & 0.90 & 96 \\
        Intrasocial   & 0.95 & 0.83 & 0.89 & 72 & 0.95 & 0.78 & 0.85 & 72 \\
        Intersocial   & 0.96 & 0.98 & 0.97 & 371 & 0.94 & 0.99 & 0.96 & 371 \\
        Intrapersonal & 0.84 & 0.94 & 0.89 & 17 & 0.76 & 0.76 & 0.76 & 17 \\
        \midrule
        \textbf{Overall}      & 0.95 & 0.95 & 0.95 & 556 & 0.93 & 0.93 & 0.93 & 556 \\
        \multicolumn{9}{c}{\textbf{Baselines: Majority = 0.67, Random = 0.49}} \\
        \bottomrule
    \end{tabular} 
    \caption{Comparison of Level Results for RoBERTa, BERT, and Baselines}
    \label{Table_Level_Comparison_with_Baselines}
\end{table*}

\begin{table*}[http]
    \centering
    \scriptsize
    \begin{tabular}{lcccc|cccc}
        \toprule
        & \multicolumn{4}{c}{\textbf{RoBERTa Results}} & \multicolumn{4}{c}{\textbf{BERT Results}} \\
        \cmidrule(lr){2-5} \cmidrule(lr){6-9}
        & \textbf{Precision} & \textbf{Recall} & \textbf{F1-Score} & \textbf{Support} 
        & \textbf{Precision} & \textbf{Recall} & \textbf{F1-Score} & \textbf{Support} \\
        \midrule
        Civilian         & 1.00 & 0.69 & 0.82 & 29 & 0.96 & 0.79 & 0.87 & 29 \\
        Jurisdictional   & 0.86 & 0.80 & 0.83 & 30 & 1.00 & 0.77 & 0.87 & 30 \\
        War/Military Campaign & 0.80 & 0.94 & 0.87 & 181 & 0.83 & 0.97 & 0.89 & 181 \\
        Battle           & 0.93 & 0.81 & 0.87 & 69 & 0.92 & 0.88 & 0.90 & 69 \\
        Plunder          & 0.69 & 0.53 & 0.60 & 17 & 0.75 & 0.53 & 0.62 & 17 \\
        Ambush           & 0.85 & 0.73 & 0.79 & 15 & 1.00 & 0.67 & 0.80 & 15 \\
        Conspiracy       & 0.82 & 0.82 & 0.82 & 11 & 0.53 & 0.82 & 0.64 & 11 \\
        Revolt           & 1.00 & 1.00 & 1.00 & 21 & 1.00 & 1.00 & 1.00 & 21 \\
        Conquest         & 0.50 & 0.57 & 0.53 & 7  & 0.57 & 0.57 & 0.57 & 7 \\
        Naval Battle     & 1.00 & 1.00 & 1.00 & 2  & 0.29 & 1.00 & 0.44 & 2 \\
        Religious        & 1.00 & 1.00 & 1.00 & 6  & 0.67 & 0.33 & 0.44 & 6 \\
        Institutional    & 0.60 & 0.75 & 0.67 & 4  & 1.00 & 0.75 & 0.86 & 4 \\
        Sack             & 0.00 & 0.00 & 0.00 & 1  & 0.00 & 0.00 & 0.00 & 1 \\
        Single Combat    & 1.00 & 0.50 & 0.67 & 4  & 1.00 & 0.50 & 0.67 & 4 \\
        Siege            & 0.83 & 0.81 & 0.82 & 31 & 0.89 & 0.81 & 0.85 & 31 \\
        Unknown          & 1.00 & 1.00 & 1.00 & 5  & 1.00 & 0.80 & 0.89 & 5 \\
        Regicide         & 0.69 & 1.00 & 0.81 & 11 & 0.79 & 1.00 & 0.88 & 11 \\
        Military         & 0.90 & 0.87 & 0.89 & 93 & 0.91 & 0.90 & 0.91 & 93 \\
        Entertaining     & 0.60 & 0.43 & 0.50 & 7  & 0.60 & 0.43 & 0.50 & 7 \\
        Mutiny           & 1.00 & 0.75 & 0.86 & 8  & 1.00 & 0.75 & 0.86 & 8 \\
        Familicide       & 1.00 & 1.00 & 1.00 & 2  & 0.00 & 0.00 & 0.00 & 2 \\
        Fratricide       & 0.00 & 0.00 & 0.00 & 1  & 0.00 & 0.00 & 0.00 & 1 \\
        Paramilitary     & 1.00 & 1.00 & 1.00 & 1  & 0.00 & 0.00 & 0.00 & 1 \\
        \midrule
        \textbf{Overall} & 0.86 & 0.85 & 0.84 & 556 & 0.86 & 0.86 & 0.85 & 556 \\
        \multicolumn{9}{c}{\textbf{Baselines: Majority = 0.33, Random = 0.16}} \\
        \bottomrule
    \end{tabular} 
    \caption{Comparison of Context Results for RoBERTa, BERT, and Baselines}
    \label{Table_Context_Comparison_with_Baselines}
\end{table*}

\begin{table*}[http]
    \centering
    \scriptsize
    \begin{tabular}{lcccc|cccc}
        \toprule
        & \multicolumn{4}{c}{\textbf{RoBERTa Results}} & \multicolumn{4}{c}{\textbf{BERT Results}} \\
        \cmidrule(lr){2-5} \cmidrule(lr){6-9}
        & \textbf{Precision} & \textbf{Recall} & \textbf{F1-Score} & \textbf{Support} 
        & \textbf{Precision} & \textbf{Recall} & \textbf{F1-Score} & \textbf{Support} \\
        \midrule
        Unknown          & 1.00 & 0.80 & 0.89 & 20 & 0.81 & 0.65 & 0.72 & 20 \\
        Political        & 0.84 & 0.86 & 0.85 & 122 & 0.91 & 0.86 & 0.89 & 122 \\
        Tactical/Strategical & 0.87 & 0.88 & 0.87 & 197 & 0.92 & 0.88 & 0.90 & 197 \\
        Economical       & 0.74 & 0.82 & 0.78 & 28 & 0.69 & 0.86 & 0.76 & 28 \\
        Following Orders & 0.90 & 0.86 & 0.88 & 77 & 0.81 & 0.90 & 0.85 & 77 \\
        Self-Defence     & 0.75 & 0.69 & 0.72 & 13 & 0.73 & 0.62 & 0.67 & 13 \\
        Emotional        & 0.97 & 0.77 & 0.86 & 43 & 0.92 & 0.84 & 0.88 & 43 \\
        Ambition         & 0.71 & 0.83 & 0.76 & 35 & 0.64 & 0.83 & 0.72 & 35 \\
        Social           & 0.71 & 1.00 & 0.83 & 5  & 1.00 & 1.00 & 1.00 & 5 \\
        Religious        & 0.83 & 0.83 & 0.83 & 6  & 0.83 & 0.83 & 0.83 & 6 \\
        Other            & 1.00 & 1.00 & 1.00 & 6  & 1.00 & 0.83 & 0.91 & 6 \\
        None/Accident    & 0.75 & 0.75 & 0.75 & 4  & 0.75 & 0.75 & 0.75 & 4 \\
        \midrule
        \textbf{Overall} & 0.86 & 0.85 & 0.85 & 556 & 0.86 & 0.86 & 0.86 & 556 \\
        \multicolumn{9}{c}{\textbf{Baselines: Majority = 0.35, Random = 0.20}} \\
        \bottomrule
    \end{tabular} 
    \caption{Comparison of Motive Results for RoBERTa, BERT, and Baselines}
    \label{Table_Motive_Comparison_with_Baselines}
\end{table*}

\begin{table*}[http]
    \centering
    \scriptsize
    \begin{tabular}{lcccc|cccc}
        \toprule
        & \multicolumn{4}{c}{\textbf{RoBERTa Results}} & \multicolumn{4}{c}{\textbf{BERT Results}} \\
        \cmidrule(lr){2-5} \cmidrule(lr){6-9}
        & \textbf{Precision} & \textbf{Recall} & \textbf{F1-Score} & \textbf{Support} 
        & \textbf{Precision} & \textbf{Recall} & \textbf{F1-Score} & \textbf{Support} \\
        \midrule
        Unknown                  & 0.78 & 0.89 & 0.83 & 199 & 0.83 & 0.91 & 0.87 & 199 \\
        Campaign                 & 0.81 & 0.87 & 0.85 & 28  & 0.82 & 0.82 & 0.82 & 28 \\
        Conquest                 & 0.83 & 0.83 & 0.83 & 24  & 0.58 & 0.92 & 0.71 & 24 \\
        Coronation/Inauguration  & 1.00 & 0.67 & 0.80 & 12  & 0.90 & 0.75 & 0.82 & 12 \\
        Exile                    & 1.00 & 0.67 & 0.80 & 6   & 0.86 & 1.00 & 0.92 & 6 \\
        Death                    & 0.81 & 0.72 & 0.72 & 32  & 0.77 & 0.69 & 0.73 & 54 \\
        Other                    & 0.72 & 0.72 & 0.72 & 32  & 0.86 & 0.78 & 0.82 & 32 \\
        Victory                  & 1.00 & 1.00 & 1.00 & 16  & 0.88 & 0.94 & 0.91 & 16 \\
        Bestowing of Honors      & 0.67 & 0.33 & 0.44 & 6   & 1.00 & 0.17 & 0.29 & 6 \\
        Issuing of Law/Decrees   & 1.00 & 0.33 & 0.50 & 3   & 0.50 & 0.33 & 0.40 & 3 \\
        Injury                   & 0.71 & 1.00 & 0.83 & 5   & 1.00 & 1.00 & 1.00 & 5 \\
        Battle                   & 0.80 & 0.53 & 0.64 & 15  & 0.67 & 0.67 & 0.67 & 15 \\
        Declaration of War       & 1.00 & 1.00 & 1.00 & 2   & 1.00 & 1.00 & 1.00 & 2 \\
        Retreat                  & 0.67 & 0.80 & 0.73 & 10  & 0.67 & 0.80 & 0.73 & 10 \\
        Mutiny                   & 1.00 & 0.50 & 0.67 & 2   & 1.00 & 0.50 & 0.67 & 2 \\
        Sending of Envoys        & 0.93 & 1.00 & 0.96 & 13  & 0.92 & 0.92 & 0.92 & 13 \\
        Civil Conflict/Civil War & 0.00 & 0.00 & 0.00 & 1   & 0.00 & 0.00 & 0.00 & 1 \\
        Tyranny                  & 0.50 & 1.00 & 0.67 & 2   & 1.00 & 1.00 & 1.00 & 2 \\
        Capture                  & 0.71 & 0.71 & 0.71 & 14  & 0.77 & 0.71 & 0.74 & 14 \\
        Destruction/Devastation  & 0.84 & 0.81 & 0.82 & 26  & 0.84 & 0.81 & 0.82 & 26 \\
        Repopulation             & 1.00 & 1.00 & 1.00 & 2   & 1.00 & 1.00 & 1.00 & 2 \\
        Declaration of Peace/Truce & 1.00 & 0.44 & 0.62 & 9 & 1.00 & 0.44 & 0.62 & 9 \\
        Release of Prisoners     & 1.00 & 1.00 & 1.00 & 2   & 0.67 & 1.00 & 0.80 & 2 \\
        Garrisoning of Troops    & 1.00 & 0.67 & 0.80 & 6   & 1.00 & 0.67 & 0.80 & 6 \\
        Famine                   & 1.00 & 1.00 & 1.00 & 1   & 1.00 & 1.00 & 1.00 & 1 \\
        Siege                    & 0.95 & 0.70 & 0.81 & 30  & 0.95 & 0.70 & 0.81 & 30 \\
        Deportation              & 1.00 & 0.25 & 0.40 & 4   & 1.00 & 0.50 & 0.67 & 4 \\
        Treaty/Agreement/Pact    & 1.00 & 0.33 & 0.50 & 3   & 0.00 & 0.00 & 0.00 & 3 \\
        Surrender                & 0.67 & 1.00 & 0.80 & 2   & 0.67 & 1.00 & 0.80 & 2 \\
        Financial Reward         & 0.75 & 1.00 & 0.86 & 3   & 0.75 & 1.00 & 0.86 & 3 \\
        Seclusion                & 0.33 & 1.00 & 0.50 & 2   & 1.00 & 1.00 & 1.00 & 2 \\
        Plunder                  & 0.86 & 1.00 & 0.92 & 6   & 1.00 & 1.00 & 1.00 & 6 \\
        Mutilation               & 1.00 & 1.00 & 1.00 & 1   & 1.00 & 1.00 & 1.00 & 1 \\
        Revenge                  & 1.00 & 1.00 & 1.00 & 6   & 1.00 & 1.00 & 1.00 & 6 \\
        Execution                & 0.40 & 0.50 & 0.44 & 4   & 0.33 & 0.25 & 0.29 & 4 \\
        Torture                  & 0.75 & 1.00 & 0.86 & 3   & 0.75 & 1.00 & 0.86 & 3 \\
        Applause                 & 1.00 & 0.50 & 0.67 & 2   & 1.00 & 0.50 & 0.67 & 2 \\
        \midrule
        \textbf{Overall}         & 0.82 & 0.80 & 0.80 & 556 & 0.82 & 0.81 & 0.81 & 556 \\
        \multicolumn{9}{c}{\textbf{Baselines: Majority = 0.36, Random = 0.16}} \\
        \bottomrule
    \end{tabular} 
    \caption{Comparison of Long-Term Consequences Results for RoBERTa, BERT, and Baselines}
    \label{Table_Long_Term_Comparison_with_Baselines}
\end{table*}

\normalsize 
\clearpage
\end{document}